\pdfoutput=1
\documentclass{article}

\usepackage[preprint]{neurips_2021}

\usepackage[utf8]{inputenc} 
\usepackage[T1]{fontenc}    
\usepackage{hyperref}       
\usepackage{url}            
\usepackage{booktabs}       
\usepackage{amsfonts}       
\usepackage{amsmath}        
\usepackage{nicefrac}       
\usepackage{microtype}      
\usepackage{xcolor}         
\usepackage{graphicx}
\usepackage{subcaption}
\usepackage{multirow}
\usepackage{geometry}

\newcommand{\bb}{\mathbf{b}}
\newcommand{\xx}{\mathbf{x}}
\newcommand{\uu}{\mathbf{u}}
\newcommand{\ff}{\mathbf{f}}

\hypersetup{
	pdftitle={StAnD: A Dataset of Linear Static Analysis Problems},    
	pdfauthor={Luca Grementieri, Francesco Finalli},     
	pdfcreator={Luca Grementieri},   
	pdfproducer={Luca Grementieri}, 
	pdfkeywords={Linear Static Analysis} {Dataset} {Computational Benchmark} {Sparse Linear Solvers} {Frame Structures}, 
	pdfnewwindow=true,      
	colorlinks=false,       
	linkcolor=red,          
	citecolor=green,        
	filecolor=magenta,      
	urlcolor=cyan           
}

\title{StAnD: A Dataset of Linear Static Analysis Problems}

\author{%
	Luca~Grementieri \\
	Zuru Tech\\
	\texttt{luca.g@zuru.tech} \\
	\And
	Francesco~Finelli \\
	Zuru Tech\\
	\texttt{francesco.f@zuru.tech} \\
}

\begin{document}
	
	\maketitle
	
	\begin{abstract}
		Static analysis of structures is a fundamental step for determining the stability of structures. Both linear and non-linear static analyses consist of the resolution of sparse linear systems obtained by the finite element method.
		The development of fast and optimized solvers for sparse linear systems appearing in structural engineering requires data to compare existing approaches, tune algorithms or to evaluate new ideas.
		We introduce the Static Analysis Dataset (StAnD) containing 303.000 static analysis problems obtained applying realistic loads to simulated frame structures. 
		Along with the dataset, we publish a detailed benchmark comparison of the running time of existing solvers both on CPU and GPU. We release the code used to generate the dataset and benchmark existing solvers on Github.
		To the best of our knowledge, this is the largest dataset for static analysis problems and it is the first public dataset of sparse linear systems (containing both the matrix and a realistic constant term).
	\end{abstract}
	
	\section{Introduction}
	Structural analysis represents the set of theories and methods which allow to idealize a structure through a mathematical model able to predict their response in terms of displacements and internal stresses, when subject to various types of external actions. The determination of response quantities plays a crucial role for the design of safe and robust constructions in civil engineering applications.
	
	The structural behavior of a solid body is analytically described by partial differential equations derived from the continuum mechanics principles of equilibrium and compatibility, related through the material constitutive relations.
	Unfortunately, analytical models of structures rarely have closed form solutions in practical engineering problems.
	Consequently, the continuum system is usually approximated with a discrete system with finite number of degrees of freedom, while the set of governing differential equations is converted and approximated with a system of algebraic equations, solvable with a variety of numerical methods.
	
	In this context, the Finite Element Method represents the most widespread technique for numerical analysis of structures. The method produces a discrete idealization of the structure which in static domain leads to a sparse algebraic system
	\begin{equation*}
		K \uu = \mathbf{f}
	\end{equation*}
	where the variable term \textit{$\uu$}, the constant term $\mathbf{f}$, and the coefficient matrix \textit{K} respectively represent the displacement vector, the external load vector, and the global stiffness matrix.
	
	The solution of such algebraic system provides the values of generalized displacements at each degree of freedom of the structure consequent to the prescribed loading conditions. Once displacements are known, then strains and stresses can be computed following kinematic compatibility conditions and material constitutive law, so that all response quantities necessary for structural design procedures are determined.
	
	
	
	Many algorithms for solving sparse linear systems are published at a great pace, but currently there is no standard dataset to compare their running time on real problems.
	
	We introduce Static Analysis Dataset (StAnD) to formalize the evaluation of new resolution methods and to spur research in resolution methods specifically tailored to structural engineering and static analysis problems.
	
	To best of our knowledge there is no existing large dataset of sparse linear systems. 
	A few sparse matrices (often less than 10) are published for many engineering problems in the Matrix Market \citep{matrixmarket} or in the SuiteSparse matrix collection \citep{suitesparsematrix}, but their limited number is not sufficient to measure the running time in the average case or to measure reliably how the resolution algorithm scales with the size and the number of non-zeros in the sparse matrix. Additionally, 
	no constant term is provided in conjunction with the matrices. The constant term, is maybe not fundamental when direct methods are used, but it becomes important if we want to measure the effectiveness of iterative methods, whose behavior depends on the relationship between the initialization of the solution and the real solution.
	
	A large-scale dataset can also pave the way to novel methods based on machine learning. Large-scale datasets have led to incredible advances in
	many areas of artificial intelligence in the last few years, like computer vision and natural language processing and we hope to foster research in machine learning application for structural engineering releasing this dataset.
	For this reason StAnD is already divided into a training split and a test split.
	
	In summary, our contribution is threefold:
	\begin{itemize}
		\item We publish a novel dataset of static analysis problems of frame structures.
		The dataset is composed of a training set of 300.000 problems and 3.000 test problems. Every split is evenly divided into small,\footnote{\url{https://storage.googleapis.com/zurutech-public-datasets/stand/stand_small.zip}}
		medium\footnote{\url{https://storage.googleapis.com/zurutech-public-datasets/stand/stand_medium.zip}}
		and large problems.\footnote{\url{https://storage.googleapis.com/zurutech-public-datasets/stand/stand_large.zip}}
		This clear separation allow to inspect the scaling behavior of solvers.
		The structures are grid-like structures procedurally designed, while the loads applied are obtained simulating elementary actions of permanent (proper weights of the structure and of non-structural elements) and variable loads (furniture, people, wind, snow).
		\item We conduct a thorough evaluation of existing open-source implementations of direct and iterative methods for the resolution of sparse linear systems.
		In particular, we put relevant attention into implementations that support GPUs since GPUs are quite common nowadays, but many algorithms are not really suited to leverage their processing power at its best.
		\item We release the code to generate the dataset based on OpenSeesPy \citep{zhu2018openseespy} and we publish the code and the Docker container used to benchmark existing solvers.\footnote{\url{https://github.com/zurutech/stand}}
	\end{itemize}
	
	\section{Related Work}
	
	\subsection{Existing Datasets}
	One the earliest collection of sparse matrices is the Matrix Market \citep{matrixmarket}, which includes in particular the Harwell-Boeing Sparse Matrix Collection \citep{duff1989sparse} and SPARSEKIT collection \citep{saad1994sparskit}.
	At the moment, the largest and most comprehensive collection of sparse matrices publicly available is the SuiteSparse Matrix collection \citep{suitesparsematrix}, which includes also a large subset of the Matrix Market.
	
	These collections are generic with sparse matrices from several domains like aerodynamics, fluid dynamics, structural engineering and economics. Additionally they are very small: in fact the SuiteSparse Matrix collection is composed of less than 3000 matrices. For every single application, they contain only a handful of matrices and it is difficult to understand if the available data faithfully reproduces the setting of interest because no metadata about the problem that has led to such matrices exists.
	These datasets can be used to compare generic solvers across several domains, but they are not so useful to compare algorithms for a specific domain of application.
	Furthermore, both collections only contain matrices and if we want to test an algorithm for the resolution of linear system, we have to sample a random vector of constant terms. The sampled constant terms can be physically unrealistic or infeasible and their choice can influence the evaluation of iterative methods.
	
	\subsection{Methods}
	
	The number of methods for the resolution of a sparse linear system $A \xx = \bb$ is large and many approaches have been suggested to tackle this classical problem.
	Linear systems for the static analysis of structures are always symmetric positive definite, thus we focus on algorithms applicable in this setting.
	The methods can be classified into three groups: direct, iterative and hybrid methods.
	We describe here the methods implemented in the most used open source libraries because these are the methods included in our benchmark.
	
	\subsubsection{Direct Methods}
	\citet{davis_rajamanickam_sid-lakhdar_2016} give a comprehensive review of direct methods and their historical development. Most direct method work in three stages: matrix permutation, matrix factorization, and triangular system resolution.
	
	\textbf{Matrix permutation.} A sparse matrix must typically be permuted either before or during its numeric factorization, either for reducing fill-in or for numerical stability. \emph{Fill-in} is the introduction of new non-zeros in the factors that
	do not appear in the corresponding positions in the matrix being factorized.
	Finding the permutation that minimizes the number of fill-in is a NP-hard problem \citep{rose1978algorithmic, yannakakis1981computing, luce2014minimum}, so the literature propose many heuristic fill-reducing methods.
	The libraries included in our benchmark implement the following permutation algorithms:
	symmetric reverse Cuthill-McKee ordering (RCM) \citep{cuthill1969reducing}, minimum degree ordering (MMD) \citep{george1989evolution}, symmetric approximate minimum degree ordering (AMD) \citep{amestoy1996approximate}, column approximate minimum degree ordering (COLAMD) \citep{davis2004column}, METIS \citep{karypis1998fast} and its parallel variant ParMETIS \citep{lasalle2016parallel}. 
	
	\textbf{Matrix factorization.} There are three main families of factorizations used for sparse linear system resolution: QR factorization ($A = QR$, with $Q$ an orthogonal matrix and $R$ an upper triangular matrix), LU factorization ($A = LU$, with $L$ a lower triangular matrix and $U$ an upper triangular matrix) and, if $A$ is symmetric positive-definite, Cholesky factorization ($A = LL^T$, with $L$ a lower triangular matrix). 
	Several specialized algorithms exist to perform such decompositions efficiently with sparse matrices. In almost all methods, the factorization splits into
	two phases: a symbolic phase that typically depends only on the non-zero
	pattern of $A$, and a numerical phase that produces the factorization itself.
	The symbolic phase finds the non-zero pattern of the triangular factors of $A$ without determining the values of the triangular factors itself.
	This phase improves the efficiency of the subsequent numerical phase and its result can be reused on matrices with identical non-zero pattern, a common situation when solving non-linear or differential equations.
	The numerical phase can be implemented in many different variants, even parallel ones. We will refer to the specific implementations in section \ref{sec:libraries_direct_methods}.
	
	\textbf{Triangular system resolution.} After the factorization of the sparse matrix $A$, we can recover the solution solving one or two consecutive triangular systems. 
	Using the QR factorization, the triangular system to be solved is
	\begin{equation*}
		R \xx = \mathbf{y} \quad \text{where} \quad y = Q^T\bb,
	\end{equation*}
	while for the LU factorization (and similarly for Cholesky factorization) the final solution is obtained solving two consecutive triangular systems
	\begin{equation*}
		L \mathbf{y} = \bb \quad \text{then} \quad U \xx = \mathbf{y}.
	\end{equation*}
	The algorithm to solve (sparse) triangular systems is quite similar to the matrix-vector multiplication algorithm, but it is inherently sequential. This implies that it is efficient like matrix-vector multiplication on CPUs, but on GPUs it is much slower.
	
	\subsubsection{Iterative Methods}
	Iterative methods are increasingly popular for the resolution of sparse linear systems.
	Here we list only methods implemented in the open-source libraries of our benchmark. Please refer to the book by \citet{saad2003iterative} for a detailed and complete introduction.
	Iterative methods are the workhorse of modern hybrid methods, so we are going to introduce them together. Hybrid methods apply an existing iterative algorithm to a preconditioned sparse system. The preconditioning transforms the linear system in an equivalent one, while reducing the condition number of the sparse matrix.
	Since the condition number determines the rate of convergence of iterative algorithms,
	this preliminary step can greatly reduce the number of iterations needed and consequently the running time of the algorithm.
	
	\textbf{Krylov methods.} The most successful iterative methods are the conjugate gradient  (CG) for symmetric positive-definite matrices and the generalized minimal residual method (GMRES) \citep{saad1986gmres} for non-symmetric matrices. Both methods fall into the family of Krylov methods since they
	are based on projections onto Krylov subspaces. A \emph{Krylov subspace} is the subspace spanned by vectors of the form $p(A)b$, where $p$ is a polynomial. In short, these
	techniques approximate $A^{-1}\bb$ by $p(A)\bb$. Other methods included in our benchmark are flexible conjugate gradient \citep{notay2000flexible}, flexible GMRES \citep{giraud2010flexible}, quasi minimal residual \citep{chan1998transpose}, conjugate residual method \citep{hestenes1952methods}, minimal residual and symmetric LQ \citep{paige1975solution}.
	
	\textbf{Preconditioning.} A \emph{preconditioner} $P$ of a matrix $A$ is a matrix such that
	$P^{-1}A$ or $AP^{-1}$ have a smaller condition number than $A$. The \emph{left preconditioned} system is the equivalent system
	\begin{equation*}
		P^{-1}A \xx = P^{-1}\bb.
	\end{equation*}
	Instead, \emph{right preconditioning} requires the resolution of two systems:
	\begin{equation*}
		A P^{-1} \mathbf{y} = \bb \text{ then } P\xx = \mathbf{y}.
	\end{equation*}
	Most preconditioners are inspired from direct methods or classic iterative methods.
	For example, the incomplete factorization ILU(0) decomposes the matrix $A$ as 
	$A = LU - R$ where $L$ and $U$ have the same nonzero structure as the lower and
	upper parts of $A$ respectively, and $R$ is the residual or error of the factorization.
	The preconditioners covered by the libraries in our benchmark are Jacobi, SOR, SSOR, ILU(0), BlockILU(0), ICC.
	
	\section{Dataset Description}
	
	The StAnD dataset is composed of 303.000 static analysis problems of frame structures divided into 6 parts: 100.000 small training problems, 100.000 medium training problems, 100.000 large training problems, 1.000 small test problems, 1.000 medium test problems and 1.000 large training problems. The size of a problem is determined by the number of degrees of freedom (DOFs) of the structure model or equivalently by the number of rows or columns of the stiffness matrix associated to the structure.
	Small problems have 2115 DOFs on average (and at most 5166 DOFs), medium problems have about 7000 DOFs (and at most 14718 DOFs), while large problems can have up to 31770 DOFs with about 15500 DOFs on average.
	The division of problems by size allows to compare the scalability of solvers.
	
	The dataset is programmatically created using OpenSeesPy \citep{zhu2018openseespy}, the Python interface to the OpenSees finite element solver \citep{mazzoni2006opensees}. For each structural model with its loading configuration, a static analysis is performed in order to compute nodal displacements. Therefore, every problem in the dataset is a tuple $(K, \ff, \uu)$, where $K$ is the sparse stiffness matrix associated to the structure, $\ff$ is a load vector and $\uu$ is the ground-truth displacement vector such that $K \uu = \ff$.
	In the training set, for the same structure (i.e. the same stiffness matrix $K$), we apply different load configurations. In the test set, we use a single load configuration for every structure to maximize the variability.
	
	\subsection{Frame Structure Generation}
	
	\begin{figure}[htb]
		\centering
		\begin{subfigure}{.3\textwidth}
			\centering
			\includegraphics[width=\textwidth]{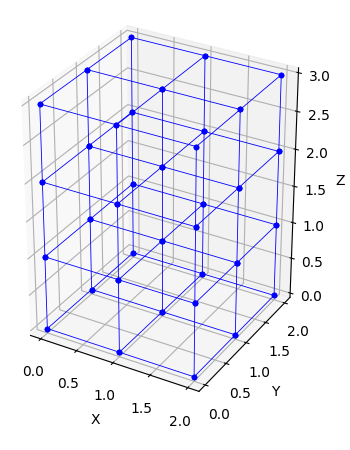}
			\subcaption{}
			\label{fig:step1}
		\end{subfigure}%
		\begin{subfigure}{.3\textwidth}
			\centering
			\includegraphics[width=\textwidth]{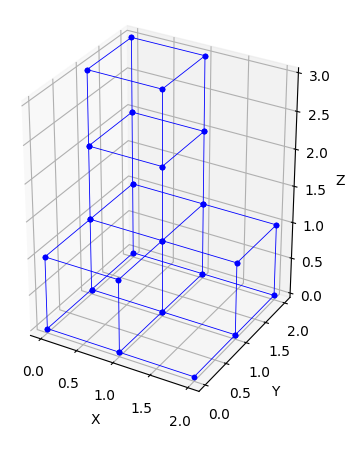}
			\subcaption{}
			\label{fig:step2}
		\end{subfigure}
		\begin{subfigure}{.3\textwidth}
			\centering
			\includegraphics[width=\textwidth]{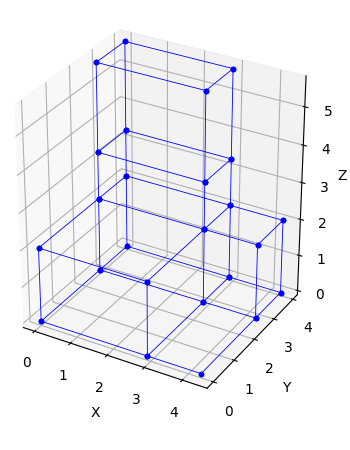}
			\subcaption{}
			\label{fig:step3}
		\end{subfigure}
		\caption{A frame structure is created from a regular 3D grid (a), removing cubes (b) and then changing the length of beams (c).}
		\label{fig:generation}
	\end{figure}
	
	Structures are created from a regular 3D grid composed of $B \times D \times H$ unit cubes. The values $B, D, H$ that determine the size of the starting grid are sampled uniformly in ranges according to the category (small, medium, or large) of the structure.
	From the regular grid, we remove a random number of cubes (from 0 to $H$ included) from each of the $b \times d$ columns made of $h$ cubes. The cubes are removed from the top, creating structures composed of stacks of cubes of different heights.
	Then the carved grid and its cubes are deformed: the thickness of every plane of cubes is changed from 1m to a random value between 3 and 6 meters. This applies to horizontal planes of cubes orthogonal to the $z$ axis (i.e. with cubes at the same level of height) and to vertical planes orthogonal to the $x$ or the $y$ axis.
	This deformation transforms the cubes into rectangular parallelepipeds.
	Finally, the frame structure is created from the edges of the deformed grid.
	The deformation of the grid is instrumental to vary the lengths of beams.
	
	The beams elements used follow the Timoshenko beam model. The parameters needed to fully characterize the beam model are sampled from the ranges shown in Table \ref{tab:beam} and they are shared by all beams in the same structure.

	\begin{table}[htb]
		\caption{Beam parameters ranges}
		\label{tab:beam}
		\centering
		\begin{tabular}{ccccc}
			\toprule
			Length (m) & Width (m) & Height (m) & Young's modulus (Pa) & Density (kg/$\text{m}^3$) \\
			\midrule
			$[3.0 , 6.0]$ & $[0.2, 0.4]$ & $[0.4, 0.6]$ & $[2.5 \cdot 10^{10}, 4.3 \cdot 10^{10}]$ & $[2.0 \cdot 10^3, 2.6 \cdot 10^3]$ \\
			\bottomrule
		\end{tabular}
	\end{table}
	
	\subsection{Load Application}
	
	The verification of a structure consists in its verification against ultimate limit state (ULS) and serviceability limit state (SLS). The loading conditions at limit states are obtained defining various combinations of several elementary loads. 
	The elementary actions are the ones induced by various sources of gravity loads, such as permanent structural and non-structural dead weights or variable live loads, by the atmospheric actions, such as wind pressure, or by earthquakes.
	In linear static analysis, we do not need to solve a sparse linear system for every combination of such actions, because the principle of superposition allows to solve the algebraic system associated to each elementary condition separately and then linearly combine the solutions.
	For example, if a structure's ULS is obtained as the combination of the dead weight of the structure, a variable live load and the action of wind, then the load vector is
	\begin{equation*}
		\ff = \varphi_g \ff_g + \varphi_v \ff_v + \varphi_w \ff_w.
	\end{equation*}
	In this case, the displacement vector $\uu$ such that $K \uu = \ff$ is
	\begin{equation*}
		\uu = \varphi_g \uu_g + \varphi_v \uu_v + \varphi_w \uu_w \text{ where }
		K\uu_g = \ff_g, K\uu_v = \ff_v, K\uu_w = \ff_w.
	\end{equation*}
	Following the engineering practice, in StAnD we include only problems obtained applying a single elementary action. The considered load typologies have been evaluated and computed according to the prescriptions of the Italian building code \citep{NTC2018}.
	
	\textbf{Proper weight of the structure.} For every structural member in StAnD we sample a density $\rho$, a section width $S_w$ and a section height $S_h$ from the ranges specified in Table \ref{tab:beam}. We denote with $g$ the standard gravitational acceleration for the surface of the Earth. For beams the proper weight load is a uniform load directed towards the negative direction of the $z$ axis with magnitude
	\begin{equation*}
		p_g = g \cdot \rho \cdot S_w \cdot S_h.
	\end{equation*}
	
	\textbf{Variable vertical pressure.} In addition to its proper weight, the structure has to support the weight of slabs and additional variable loads linked to the function of the structure itself. Slabs could be modelled more precisely with a 4-node element, but since we are working with frame structures only, we have to compute the nodal equivalent loads.
	We assume that every slab transfers all loads to the horizontal beams just below it.
	Under this simplifying assumption, a uniform load on a rectangular slab is equivalent to a pair of identical triangle loads on the shortest sides of the rectangle and a pair of trapezoidal loads on the longest sides of the rectangle, as shown in Figure
	In practice, triangle and trapezoidal loads are further simplified into uniform loads with the same total force on the beam.
	In conclusion, denoting $\ell$ and $L$ the short and long dimensions of the beams of the slab, the uniform loads on beams of the slab have magnitude
	\begin{equation*}
		p_v^{\text{short}} = P \cdot \frac{\ell}{4} \qquad
		p_v^{\text{long}} = P \cdot \frac{\ell}{2L} \cdot \left(L - \frac{\ell}{2}\right),
	\end{equation*}
	where $P$ is the vertical pressure on the slab.
	
	\textbf{Wind pressure.} The action of the wind is modelled as pressures acting on the windward face of the structure and depressions on the leeward and lateral faces.
	The direction of the wind in static analysis is taken parallel to the principal axes of the
	building plan. The pressure of the wind depends on the altitude $z$ and it is given by
	\begin{equation*}
		p_w(z) = \frac{1}{2} \rho_{\text{air}} \cdot v^2 \cdot c_e(z) \cdot c_p,
	\end{equation*}
	where $\rho_{air} = 1.25 \text{kg/m}^3$ is the density of the air, $v$ is the speed of wind,
	$c_e(z)$ is the coefficient of exposition (which is a non-decreasing function of $z$) and
	$c_p$ is the pressure coefficient of the face under examination.
	The coefficient of exposition is defined as
	\begin{equation*}
		c_e(z) =   
		\begin{cases}
			\displaystyle k^2 \ln\left(\frac{z}{z_0} \right)\left[ 7 + \ln\left(\frac{z}{z_0}\right) \right] & \text{ if } z \geq z_{\text{min}} \\
			c_e(z_{min}) & \text{ if } z < z_{\text{min}}.
		\end{cases}
	\end{equation*}
	The parameters $k$, $z_0$ and $z_{\text{min}}$ depend on the category of exposition of the building site and they are chosen according to Table \ref{tab:exposition}.
	\begin{table}[htb]
		\caption{Parameters determining the coefficient of exposition}
		\label{tab:exposition}
		\centering
		\begin{tabular}{cccc}
			\toprule
			Category of exposition & $k$ & $z_0$ (m) & $z_{\text{min}}$ (m) \\
			\midrule
			I & $0.17$ & $0.01$ & $2$ \\
			II & $0.19$ & $0.05$ & $4$ \\
			III & $0.20$ & $0.10$ & $5$ \\
			IV & $0.22$ & $0.30$ & $8$ \\
			V & $0.23$ & $0.70$ & $12$ \\
			\bottomrule
		\end{tabular}
	\end{table}
	The pressure coefficient depends on the ratio between the height of the structure $h$ and its depth $d$ measured along the axis parallel to the direction of the wind.
	The positive sign of the pressure coefficient indicates a pressure, while the negative sign
	signal a depression.
	The pressure coefficient has different formulations for the windward, the leeward and the lateral faces of the structure:
	\begin{equation*}
		c_p^{\text{windward}}(h, d) =
		\begin{cases}
			0.7 + 0.1 \frac{h}{d} & \text{ if } \frac{h}{d} \leq 1 \\
			0.8 & \text{ if } \frac{h}{d} > 1;
		\end{cases}
	\end{equation*}
	\begin{equation*}
		c_p^{\text{leeward}}(h, d) =
		\begin{cases}
			-0.3 - 0.2 \frac{h}{d} & \text{ if } \frac{h}{d} \leq 1 \\
			-0.5 - 0.05 (\frac{h}{d}-1) & \text{ if } \frac{h}{d} > 1;
		\end{cases}
	\end{equation*}
	\begin{equation*}
		c_p^{\text{lateral}}(h, d) =
		\begin{cases}
			-0.5 - 0.8 \frac{h}{d} & \text{ if } \frac{h}{d} \leq \frac{1}{2} \\
			-0.9 & \text{ if } \frac{h}{d} > \frac{1}{2}.
		\end{cases}
	\end{equation*}
	
	We suppose that every external frame of structure (composed of a pair of horizontal beams and a pair of vertical columns) is covered by a wall able to transfer all the pressure of the wind on beams and columns. The load on a wall is not uniform since it varies with the altitude, but for simplicity we fix the value of the coefficient of exposition to $c_e(z^{\text{top}})$ fro the entire loaded surface obtaining a uniform load. Since $c_e(z)$ is a non-decreasing function and $z^{\text{top}}$ is the highest altitude of the wall, the resulting load is an upper bound over the real load.
	Lastly, the wind uniform load is distributed on beams using the same approximation introduced for vertical pressure on slabs.
	
	\textbf{Snow pressure.} The pressure of the snow on the roof of the structure is modelled similarly to a variable vertical pressure, but it can only be applied to slabs at the top of the structure. The formula modelling the snow pressure is
	\begin{equation*}
		p_s = p_0 \cdot \mu,
	\end{equation*}
	where $p_0$ is the load of the snow on the ground and $\mu$ is a coefficient that depends on the slope of the roof.
	The value $p_0$ is sampled in the interval $[600.0, 5600.0]$ to cover different weather conditions. The coefficient $\mu = 0.8$ is constant for all structures in the dataset since
	the roof is always flat.
	
	\section{Benchmark}
	
	We use StAnD to measure the running time of linear solvers on static analysis problem.
	Our benchmark try to be one of the most comprehensive, with a particular focus on libraries with GPU support. We will see that not all algorithms are faster on GPUs than on CPUs as we would expect. GPUs shines for computations with a high degree of parallelization and low degree of inter-communication. If the algorithm is inherently sequential and it does not satisfy thee requirements, then the CPU's running time can be lower than the GPU's running time.
	
	\subsection{Open-source Libraries}
	\label{sec:libraries_direct_methods}
	Our benchmark includes the following open-source C++ libraries with GPU support:
	SuiteSparse \citep{davis2019algorithm}, cuSOLVER \citep{cusolver}, SuperLU\_DIST \citep{lidemmel03}, PETSc \citep{petsc-user-ref} and ViennaCL \citep{rupp2016viennacl}.
	SuiteSparse, cuSOLVER and SuperLU\_DIST implements direct methods, while PETSc and ViennaCL are focused on iterative methods.
	Since PETSc and SuiteSparse do not fully support GPUs and they have better performances on CPUs, we report only their results obtained on CPUs.
	
	\subsection{Results}
	We test the libraries on a PC equipped with 12 CPU cores (Intel i7-6850K CPU @ 3.60GHz) and a GPU Nvidia GeForce GTX 1080 Ti. 
	We measure the time needed to solve the linear system, without considering the time spent to convert the system in the format accepted by each library or the time needed to transfer data on the GPU.
	For iterative algorithms we stop the iteration when the relative error goes below $10^{-6}$ or when we reach 1000 iterations.
	Since we known the OpenSees solution $x$, we can also evaluate the \emph{standard error} of the approximation $\widetilde{x}$ with respect to $x$ defined as
	\begin{equation*}
		SE(\widetilde{x}, x) = \frac{||x-\widetilde{x}||}{||x||}.
	\end{equation*}
	We consider that an algorithm has converged for all problems if the average standard error is below $10^{-3}$.
	When an algorithm does not converge, its running time is not reported.
	
	Tables \ref{tab:cpu_results} and \ref{tab:gpu_results} show the average running time of solvers for the test split of StAnD on small, medium and large problems. As expected, the methods tailored to solve SPD systems (Cholesky factorization, conjugate gradient and conjugate residuals) are faster than methods specifically studied to solve non-SPD systems (LU and QR factorizations, GMRES, QMR, Minimum Residual and SymmLQ).
	We can notice that iterative methods are much faster than concurrent methods,
	and preconditioning definitely helps convergence.
	In particular incomplete factorization methods like ICC and ILU(0) gives a relevant speed-up on CPU (from 4x to 25x).
	On GPU, preconditioning reduces the average standard error but it often results in slower running times, probably because the factorization cannot be easily parallelized. The same difficulty in parallelization explains why direct algorithms are always faster on CPU than on GPU.
	
	In conclusion, we advice practitioners to use the (preconditioned) conjugate gradient algorithm to solve linear static analysis problems.
	Comparing the conjugate gradient algorithm implementations on CPU (PETSc) and on GPU (ViennaCL), we see that the GPU is noticeably faster only on large problems: in fact, for small and medium problems the GPU is severely underutilized.
	For this reason, we recommend to use PETSc implementation with ILU(0) preconditioning for analyses with less than 7000 DOFs and to optionally switch to ViennaCL implementation on GPU for larger structures. 
	
	\clearpage
	
	\newgeometry{top=2.2cm, bottom=2.2cm}
	
	\begin{table}[htb!]
		\caption{Benchmark timing results on CPU.}
		\label{tab:cpu_results}
		\centering
		\begin{tabular}{cccccc}
			\toprule
			\multirow{3}{*}{Library} & \multirow{3}{*}{Method} & \multirow{3}{*}{\shortstack{Permutation / \\Preconditioner}} & \multirow{3}{*}{\shortstack{Small\\problems\\time (ms)}} &
			\multirow{3}{*}{\shortstack{Medium\\problems\\time (ms)}} &
			\multirow{3}{*}{\shortstack{Large\\problems\\time (ms)}}\\
			&&&&&\\
			&&&&&\\
			\midrule
			\multirow{4}{*}{\shortstack{SuiteSparse\\(CPU)}} & KLU & - & 44.8 & 600 & 3925 \\
			& LU & - & 26.2 & 127 & 420 \\
			\cmidrule{2-6}
			& \multirow{2}{*}{Cholesky} & Simplicial & 21.9 & 227 & 896 \\
			& & Supernodal & 13.4 & 64.6 & 239 \\
			\midrule
			\multirow{45}{*}{\shortstack{PETSc\\(CPU)}} & \multirow{6}{*}{\shortstack{Conjugate\\Gradient}} & - & 11.7 & 73.4 & 253 \\
			&& Jacobi & 9.35 & 63.1 & 229 \\
			&& SOR & 8.15 & 51.9 & 180 \\
			&& SSOR & 5.87 & 38.1 & 124 \\
			&& ICC & 2.28 & 17.0 & 53.1 \\
			&& \textbf{ILU(0)} & \textbf{2.13} & 15.8 & 48.6\\
			\cmidrule{2-6}
			&\multirow{6}{*}{\shortstack{Conjugate\\Residuals}} & - & 11.0 & 67.6 & 239 \\
			&& Jacobi & 8.30 & 53.9 & 194\\
			&& SOR & 7.55 & 46.7 & 159 \\
			&& SSOR & 5.70 & 35.6 & 113 \\
			&& ICC & 2.33 & 16.8 & 51.8 \\
			&& \textbf{ILU(0)} & 2.18 & \textbf{15.7} & \textbf{47.9}\\
			\cmidrule{2-6}
			&\multirow{6}{*}{GMRES} & - & 74.7 & 149 & 442\\
			&& Jacobi & 32.1 & 141 & 440 \\
			&& SOR & 16.7 & 105 & 407 \\
			&& SSOR & 13.9 & 85.1 & 305 \\
			&& ICC & 3.44 & 23.1 & 75.5 \\
			&& ILU(0) & 3.25 & 21.9 & 71.0 \\
			\cmidrule{2-6}
			&\multirow{6}{*}{\shortstack{Flexible\\GMRES}} & - & 74.2 & 149 & 443 \\
			&& Jacobi & 38.3 & 146 & 447 \\
			&& SOR & 16.8 & 105 & 407 \\
			&& SSOR & 12.2 & 76.7 & 280 \\
			&& ICC & 3.13 & 21.2 & 69.8 \\
			&& ILU(0) & 2.96 & 20.2 & 66.0 \\
			\cmidrule{2-6}
			&\multirow{6}{*}{\shortstack{Minimum\\Residual}} & - & 14.4 & 88.2 & 312 \\
			&& Jacobi & 11.7 & 74.9 & 271 \\
			&& SOR & 8.95 & 55.3 & 191 \\
			&& SSOR & 6.81 & 42.2 & 138 \\
			&& ICC & 2.53 & 18.1 & 57.1 \\
			&& ILU(0) & 2.38 & 17.1 & 53.3 \\
			\cmidrule{2-6}
			&\multirow{6}{*}{QMR} & - & 142 & 254 & 876\\
			&& Jacobi & - & - & - \\
			&& SOR & 63.6 & 201 & 740 \\
			&& SSOR & 54.7 & 155 & 529 \\
			&& ICC & 12.0 & 62.6 & 233 \\
			&& ILU(0) & 11.9 & 57.9 & 212\\
			\cmidrule{2-6}
			&\multirow{6}{*}{SymmLQ} & - & 16.0 & 99.4 & 338 \\
			&& Jacobi & - & - & -\\
			&& SOR & - & - & -\\
			&& SSOR & - & - & -\\
			&& ICC & - & - & -\\
			&& ILU(0) & - & - & - \\
			\cmidrule{2-6}
			& Cholesky & - & 63.7 & 721 & 3526 \\
			\cmidrule{2-6}
			& LU & - & 44.3 & 490 & 2811 \\
			\cmidrule{2-6}
			& QR & - & 65.5 & 331 & 1110 \\
			\bottomrule
		\end{tabular}
	\end{table}
	
	\clearpage
	
	\restoregeometry
	
	\begin{table}[htb!]
		\caption{Benchmark timing results on GPU.}
		\label{tab:gpu_results}
		\centering
		\begin{tabular}{cccccc}
			\toprule
			\multirow{3}{*}{Library} & \multirow{3}{*}{Method} & \multirow{3}{*}{\shortstack{Permutation / \\Preconditioner}} & \multirow{3}{*}{\shortstack{Small\\problems\\time (ms)}} &
			\multirow{3}{*}{\shortstack{Medium\\problems\\time (ms)}} &
			\multirow{3}{*}{\shortstack{Large\\problems\\time (ms)}}\\
			&&&&&\\
			&&&&&\\
			\midrule
			\multirow{8}{*}{\shortstack{cuSOLVER\\(GPU)}} & \multirow{4}{*}{Cholesky} & - & 40.4 & 310 & 1339\\
			& & METIS & 30.7 & 136 & 393 \\
			& & SymAMD & 35.1 & 212 & 817 \\
			& & SymRCM & 34.5 & 226 & 866 \\
			\cmidrule{2-6}
			& \multirow{4}{*}{QR} & - & 103 & 806 & 4151 \\
			& & METIS & 154 & 1521 & 7466 \\
			& & SymAMD & 212 & 2974 & 20784\\
			& & SymRCM & 89.1 & 594 & 2571 \\
			\midrule
			\multirow{5}{*}{\shortstack{SuperLU\\(GPU)}} & \multirow{5}{*}{LU} & - & 1767 & 21399 & -\\
			& & ColAMD & 418 & 775 & 1631\\
			& & METIS & 374 & 531 & 835\\
			& & \multirow{2}{*}{\shortstack{Minimum\\Degree}} & \multirow{2}{*}{425} & \multirow{2}{*}{626} & \multirow{2}{*}{906}\\
			&&&&&\\
			\midrule
			\multirow{14}{*}{\shortstack{ViennaCL\\(GPU)}} & \multirow{7}{*}{\shortstack{Conjugate\\Gradient}} & - & 10.3 & \textbf{15.0} & \textbf{23.8} \\
			& & Row Scaling & 19.3 & 30.3 & 41.7 \\
			& & Jacobi & 19.1 & 29.9 & 41.0\\
			& & Chow-Patel & 26.0 & 43.9 & 70.2 \\
			& & \textbf{ICC} & \textbf{6.26} & 19.0 & 43.5\\
			& & ILU(0) & 6.58 & 19.4 & 43.6 \\
			& & Block ILU(0) & - & - & -\\
			\cmidrule{2-6}
			& \multirow{7}{*}{GMRES} & - & - & - & - \\
			& & Row Scaling & - & - & - \\
			& & Jacobi & - & - & - \\
			& & Chow-Patel & - & - & -\\
			& & ICC & - & - & -\\
			& & ILU(0) & - & - & -\\
			& & Block ILU(0) & - & - & -\\
			\bottomrule
		\end{tabular}
	\end{table}
	
	\section{Conclusion}
	We introduce the Static Analysis Dataset (StAnD), a novel dataset of linear static analysis problems.
	The dataset gives the possibility to evaluate direct and iterative methods for the resolution of sparse symmetric
	positive-definite linear systems. Such linear systems are obtained simulating complex frame structures of various sizes
	and applying realistic loads to them.
	
	Many open source libraries with partial or complete GPU support implementing direct or iterative methods
	were thoroughly evaluated on this dataset.
	The evaluations provide a first benchmark on this dataset and show that there is still considerable room for improvement
	to leverage the parallel processing power of GPUs, since in many settings CPUs are still faster than GPUs.
	It is our hope that the proposed dataset will stimulate the
	development of new sparse linear solvers specifically tailored for GPUs and for structural analysis.
	
	\clearpage
	
	\bibliographystyle{plainnat}
	\bibliography{stand}
\end{document}